\documentclass[conference]{IEEEtran}
\IEEEoverridecommandlockouts
\usepackage[hidelinks]{hyperref}
\usepackage{cite}
\usepackage{algorithm}
\usepackage{algpseudocode}
\usepackage{amsmath,amssymb,amsfonts}
\usepackage{graphicx}
\usepackage{textcomp}
\usepackage{xcolor}
\def\BibTeX{{\rm B\kern-.05em{\sc i\kern-.025em b}\kern-.08em
    T\kern-.1667em\lower.7ex\hbox{E}\kern-.125emX}}
\usepackage{url}

\begin{document}

\bstctlcite{IEEEexample:BSTcontrol}

\title{Bio-Inspired Magnetic Navigation Using Gradient Descent With Rao–Blackwellized Particle Filtering\\
}

\author{
\IEEEauthorblockN{
Zayed Herma\textsuperscript{1,2},
Andréa Vassilev\textsuperscript{1},
Philippe Cardin\textsuperscript{2},
Jean-Michel Léger\textsuperscript{1}
}
\IEEEauthorblockA{
\textsuperscript{1}\textit{Univ. Grenoble Alpes, CEA, Leti, 38000 Grenoble, France}\\
\textsuperscript{2}\textit{Univ. Grenoble Alpes, CNRS, ISTerre, 38000 Grenoble, France}
}
}


\maketitle

\begin{abstract}
Navigation in GPS-denied environments is commonly based on magnetic maps, which are used to correct trajectory drift and enable relocalization. However, such navigation systems suffer from a strong dependence on magnetic maps that are complex, heavy, and costly to construct. Bio-inspired magnetic navigation (BIMN) does not rely on magnetic maps, making it simpler and more frugal. In this paper, we present a gradient descent BIMN method based on the Rao-Blackwellized particle filter (RBPF) for local gradient estimation. We show that magnetic navigation can be performed without magnetic maps, as gradient descent naturally attracts the system toward the destination, resulting in a simple and memory-efficient approach.
\end{abstract}


\section{Introduction}
Modern low-cost, passive navigation methods for GPS-denied environments typically rely on high-precision, pre-constructed magnetic maps. These methods use map matching, in which the measured magnetic parameters are compared with the map to estimate the most probable measurement location and reconstruct the navigation path \cite{xu_improved_2024, abdul_raouf_maght_2023,canciani_magnetic_2022,chen_improved_2022, chen_new_2018, kok_scalable_2018}.

The main challenge of this magnetic map–based approach is the requirement for a pre-constructed magnetic map. Such a map must be constructed a priori using a dedicated mapping method \cite{menzen_large-scale_2023, solin_modeling_2018}. Once constructed, the map must be stored and continuously accessed by the navigation system, which results in significant memory consumption. In addition, magnetic maps must be regularly maintained and updated to preserve localization accuracy, further increasing system complexity and operational cost.

These issues become especially pronounced in large-scale environments, where building, storing, and processing high-resolution magnetic maps is computationally intensive and expensive. As a result, the practical utility of magnetic map–based methods is generally limited to relatively small areas, making this approach unsuitable for large-scale navigation, for which constructing and  maintaining such maps
would be extremely challenging.

\begin{figure}[!t]
    \centering
    \includegraphics[width=0.9\columnwidth]{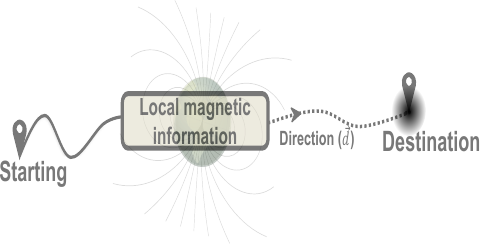}
    \caption{Illustration of BIMN. An agent navigates toward a destination defined by target magnetic parameters (e.g., magnetic field intensity, inclination, and declination of the Earth’s magnetic field). The agent has no global magnetic map and relies only on locally measured or extracted magnetic parameters, which are compared with the target values to infer the direction of movement.}
    \label{fig:illustration_bio_inspired_magnetic_navigation}
\end{figure}

This is why drawing inspiration from animals that navigate long distances using the Earth’s magnetic field could help overcome the complexity of using such magnetic maps. These animals have been demonstrated to navigate without relying on pre-constructed maps or external infrastructure, using mechanisms that are robust, efficient, and probably simple \cite{ghosh_magnetic_2024, lohmann_magnetic_2022, dennis_evidence_2007, lohmann_magnetic_2007}. For example, the long-distance migration of sea turtles across vast expanses of seemingly featureless ocean suggests that these animals rely on the Earth’s magnetic field for navigation \cite{lohmann_there_2019, walker_magnetic_2002}. It has been shown that these animals memorize and store only the magnetic parameters of their destination, not an entire magnetic map. As they travel, they continuously measure local magnetic parameters and compare them with stored destination parameters to determine the correct direction toward the destination, as illustrated in Fig.~\ref{fig:illustration_bio_inspired_magnetic_navigation} \cite{goforth_learned_2025, packmor_migratory_2024}.

This has led to the development of many bio-inspired magnetic navigation (BIMN) methods \cite{yang_bionic_2025, yang_geomagnetic-guided_2024, zhang_bio-inspired_2021, zhang_geomagnetic_2021, qi_bioinspired_2018, zhongwei_zhao_long-distance_2014}. In these studies, the gradient (representing the first-order variation of magnetic parameters) is explicitly and manually estimated to guide the agent. This estimation typically requires executing multiple maneuvers in different directions to acquire the necessary gradient information. To compensate for inaccurate or delayed gradient estimates, complex techniques (such as evolutionary optimization, predictive control, and data-driven approaches) are often employed, leading to oversized and computationally expensive solutions.

In this paper, we propose a navigation method that is fusion-based in the sense of performing online joint estimation of the local magnetic gradient and the agent’s position during navigation toward the destination. The proposed approach formulates gradient and position as coupled states and estimates them using the Rao-Blackwellized particle filter (RBPF). The estimated local gradient is then exploited to compute the direction of motion via a gradient-descent strategy. The main contributions of this paper can be summarized as follows:
\begin{itemize}
    \item Formulating the BIMN problem as a gradient-descent optimization approach (see Section~\ref{sec:problem_formulation}).
    \item Jointly estimating the local gradient and the agent’s position using the RBPF (see Section~\ref{sec:methods}).
\end{itemize}

The proposed approach is compared with a representative state-of-the-art method \cite{gill_navigation_2024}, which assumes a constant gradient. This baseline method is selected for its conceptual simplicity and bio-inspired nature. The proposed method introduces only minimal additional estimation to handle a spatially varying gradient.

The comparison highlights the limitations of constant-gradient assumptions and demonstrates the benefits of online local gradient estimation, particularly in long-range navigation scenarios where magnetic gradients vary spatially. The results show that the local gradient is estimated along the navigation path, with uncertainty that is smaller in the direction of motion, which is the direction of attraction governed by gradient descent. Consequently, estimating the gradient in all directions is unnecessary: only the component of the gradient projected onto the direction of motion matters for effective descent (see Section~\ref{sec:results}). We also show that the developed method uses the same locally estimated gradient to correct the position and attenuate drift in the integrated trajectory, as if a local map were available.

\section{PROBLEM FORMULATION}
\label{sec:problem_formulation}
In this section, we formalize the problem of BIMN as a pure optimization problem. Let
$\mathbf{B}_g = \left( B_g^{1}, \ldots, B_g^{n} \right)^{\mathsf T} \in \mathbb{R}^n$ denote the magnetic parameters of the goal destination, and
$\mathbf{B}_k = \left( B_k^{1}, \ldots, B_k^{n} \right)^{\mathsf T} \in \mathbb{R}^n$ denote the magnetic parameters measured at time step $t_k$. Here, $n$ represents the number of magnetic parameters. The problem of BIMN can be formulated as an optimization problem in which the agent minimizes the objective function

\[
J(\mathbf{B}_k) = \lVert \mathbf{B}_g - \mathbf{B}_k \rVert^2,
\]
defined as
\[
\lVert \mathbf{B}_g - \mathbf{B}_k \rVert^2
= (\mathbf{B}_g - \mathbf{B}_k)^{\mathsf T}\,\mathbf{N}^{-1}\,(\mathbf{B}_g - \mathbf{B}_k).
\]

The matrix $\mathbf{N}$ is defined as a diagonal weighting matrix,
\[
\mathbf{N} = \mathrm{diag}\!\left(\sigma_1^2, \sigma_2^2, \ldots, \sigma_n^2\right) \in \mathbb{R}^{n \times n},
\]
where each diagonal element $\sigma_i^2$ is a strictly positive scaling factor used to normalize the corresponding magnetic parameter, making its contribution to the objective function unitless (e.g., mixing an inclination angle (°) with an intensity in nanotesla (nT)). Smaller values of $\sigma_i^2$ increase the influence of that component in the optimization, while larger values decrease it.  At each iteration $k$, the agent computes a direction vector $\mathbf{d}_k$ toward the target $\mathbf{B}_g$, defined as a descent direction of the objective function.

To simplify the problem, we assume a two-dimensional navigation scenario. The navigation frame of the agent is denoted by $(\mathbf{e}_x, \mathbf{e}_y)$, and the current position of the agent in this frame is given by $\mathbf{p}_k = (x_k, y_k) \in \mathbb{R}^2$. This assumption is reasonable in outdoor navigation scenarios, where long-distance navigation can be approximated as occurring at a constant altitude, although the formulation of the problem can be easily generalized to three-dimensional case.

A simple way to iteratively minimize $J(\mathbf{B}_k)$ is the gradient descent method, in which the descent direction is given by the opposite of the gradient,
\[
-\nabla J(\mathbf{B}_k) = 2\nabla \mathbf{B}_k^{\mathsf T} \mathbf{N}^{-1} (\mathbf{B}_g - \mathbf{B}_k),
\]
corresponding to the direction of steepest decrease. This implies that the position of the agent in the navigation frame evolves between $t_k$ and $t_{k+}$ according to
\begin{equation}
\label{eq:odom_position}
\mathbf{p}_{k+1} = \mathbf{p}_{k} + v \, \Delta t \, \mathbf{d}_k ,
\end{equation}
where
\begin{itemize}
    \item $v$ is the speed of the agent,
    \item $\Delta t$ is the sampling period, and
\end{itemize}
the unit direction vector $\mathbf{d}_k$ is given by
\begin{equation}
\label{eq:normalization_d}
\mathbf{d}_k = \frac{-\nabla J(\mathbf{B}_k)}{\left\|\nabla J(\mathbf{B}_k)\right\|} = 
\frac{\nabla \mathbf{B}_{k}^{\mathsf T} \mathbf{N}^{-1} (\mathbf{B}_g - \mathbf{B}_k)}
{\left\| \nabla \mathbf{B}_{k}^{\mathsf T} \mathbf{N}^{-1} (\mathbf{B}_g - \mathbf{B}_k) \right\|},
\end{equation}

where $\nabla \mathbf{B}_k$ denotes the gradient of $\mathbf{B}_k$, defined as
\[
\nabla \mathbf{B}_k =
\begin{bmatrix}
\dfrac{\partial B_k^{1}}{\partial x} & \dfrac{\partial B_k^{1}}{\partial y} \\
\dfrac{\partial B_k^{2}}{\partial x} & \dfrac{\partial B_k^{2}}{\partial y} \\
\vdots & \vdots \\
\dfrac{\partial B_k^{n}}{\partial x} & \dfrac{\partial B_k^{n}}{\partial y}
\end{bmatrix}
\label{eq:gradient_matrix}
 \in \mathbb{R}^{n \times 2}.\]

In this case, the agent iteratively moves step by step until the norm $\|\mathbf{B}_g - \mathbf{B}_{k_0}\|$ becomes smaller than a predefined threshold $\varepsilon$ (we refer to this as the stopping threshold). At this point, the agent is considered to have reached the destination at time $t_{k_0}$. The normalization in \eqref{eq:normalization_d}, together with the fact that the velocity $v$ and the sampling period  $\Delta t$ are constant, ensures that the agent moves with a fixed step length equal to $v \, \Delta t$. This choice simplifies the navigation process by using a constant step size at each iteration, avoiding the complexity of a varying step.

The main challenge in computing $\mathbf{d}_k$ lies in accessing the gradient of the magnetic parameters $\nabla \mathbf{B}_{k}^{\mathsf T}$. The agent has no global map, and the only accessible measured information from the environment is $\mathbf{B}_{k}$. In the literature, this difficulty is commonly addressed using two main approaches: assuming a globally constant magnetic field gradient \cite{gill_navigation_2024}, or estimating the gradient using manual methods based on exploratory movements in different directions, where the gradient is approximated by the difference between successive magnetic field measurements divided by the distance between those measurements \cite{yang_bionic_2025, yang_geomagnetic-guided_2024, zhang_bio-inspired_2021, zhang_geomagnetic_2021, qi_bioinspired_2018, zhongwei_zhao_long-distance_2014}. In Section~\ref{sec:methods}, we estimate this gradient by leveraging local magnetic field measurements, the expected local variation of the gradient, and the predicted measurements from one step to the next. This estimation is performed using the RBPF, which accounts for the nonlinear navigation dynamics of the system in \eqref{eq:odom_position} and the local variation of the gradient, which is assumed to be locally linear.

\section{METHODS}
\label{sec:methods}
This section presents the dynamic and observation equations and describes the estimation method used to jointly estimate the gradient and the agent position.
\subsection{Dynamic Model of the Gradient}
\label{subsec:dynamic_model}
We denote by $\mathbf{g}_k \in \mathbb{R}^{2n}$ the flattened gradient vector of $\nabla \mathbf{B}_k$, given by
\[
\mathbf{g}_k =
\begin{bmatrix}
\dfrac{\partial B_k^{1}}{\partial x} &
\dfrac{\partial B_k^{1}}{\partial y} &
\dfrac{\partial B_k^{2}}{\partial x} &
\dfrac{\partial B_k^{2}}{\partial y} &
\cdots &
\dfrac{\partial B_k^{n}}{\partial x} &
\dfrac{\partial B_k^{n}}{\partial y}
\end{bmatrix}^{\mathsf T}.
\label{eq:flattened_gradient}
\]

As a first approximation, we assume that the gradient varies slowly with respect to the agent’s movement. We model this using the following dynamic equation:

\begin{equation}
\label{eq:dyn_gradient}
    \mathbf{g}_{k+1} = \mathbf{g}_{k} + \mathbf{q}_k
\end{equation}

where $ \mathbf{q}_k$ represents the variation of the gradient, modeled as a random variable with a normal distribution $\mathcal{N}(\mathbf{0}, \mathbf{Q}_k)$. This random variable models the fact that small variations are more likely to occur than large variations, and the direction of the variation is a priori uniformly unknown.

Equation~\eqref{eq:dyn_gradient} models the evolution of the gradient from $\mathbf{p}_k$ to $\mathbf{p}_{k+1}$. When the agent moves between steps, the uncertainty of the new gradient at time step $t_{k+1}$ increases according to the covariance matrix $\mathbf{Q}_k$ with respect to the previous gradient at time step $t_k$. In other words, the expected value of the gradient remains unchanged,
\[
    \mathbb{E}[\mathbf{g}_{k+1}] = \mathbb{E}[\mathbf{g}_k],
\]
while its covariance matrix evolves according to
\[
    \mathrm{Cov}(\mathbf{g}_{k+1}) = \mathrm{Cov}(\mathbf{g}_k) + \mathbf{Q}_k.
\]

\subsection{Observation Model}
\label{subsec:observation_model}
Since the gradient is assumed to vary slowly, the observation equation is approximated using a first-order Taylor expansion, as follows:

\begin{equation}
\label{eq:observation_v0}
\mathbf{B}_{k+1} = \mathbf{B}_{k} + \nabla \mathbf{B}_{k}\,(\mathbf{p}_{k+1} - \mathbf{p}_{k}) + \mathbf{a}_k + \mathbf{r}_{k+1}
\end{equation}

where $\mathbf{a}_k$ represents the approximation error (from the first-order expansion) and is modeled as a random variable with a normal distribution $\mathcal{N}(\mathbf{0}, \mathbf{A}_k)$. This first-order approximation is expected to locally capture the variation of the measurement. Similarly, $\mathbf{r}_{k+1}$ is modeled as a random variable with a normal distribution $\mathcal{N}(\mathbf{0}, \mathbf{R}_{k+1})$, representing the noise in the actual measurement $\mathbf{B}_{k+1}$. Equation~(\ref{eq:observation_v0}) can be rewritten as

\begin{equation}
\label{eq:observation_v1}
\mathbf{B}_{k+1}
= \mathbf{B}_{k} + \mathbf{H}_{k}\mathbf{g}_{k}
+ \mathbf{a}_{k} + \mathbf{r}_{k+1},
\end{equation}
where $\Delta x_k \triangleq x_{k+1}-x_k$ and $\Delta y_k \triangleq y_{k+1}-y_k$, and
\begin{equation}
\label{eq:Hk}
\mathbf{H}_{k}=
\left[
\begin{array}{@{}cccccc@{}}
\Delta x_k & \Delta y_k & 0          & \cdots & 0          & 0 \\
0          & 0          & \Delta x_k  & \Delta y_k & \cdots & 0 \\
\vdots     & \vdots     & \ddots      & \ddots & \ddots     & \vdots \\
0          & 0          & 0           & \cdots & \Delta x_k  & \Delta y_k
\end{array}
\right]
\in \mathbb{R}^{n\times 2n}.
\end{equation}

The equations~(\ref{eq:dyn_gradient}) and~(\ref{eq:observation_v1}) are used to continuously estimate the gradient (denoted as $\tilde{\mathbf{g}}_k \in \mathbb{R}^{2n}$ or $\tilde{\mathbf{G}}_k \in \mathbb{R}^{n \times 2}$). This estimate is then used to perform gradient descent and control the agent in minimizing the objective function $\|\mathbf{B}_g - \mathbf{B}_k\|^2$.

\subsection{Agent Motion Model}
\label{subsec:gradient_descent_dynamic_model}
As discussed in Section~\ref{sec:problem_formulation}, the agent’s motion is controlled by the gradient descent, as given in \eqref{eq:odom_position}. However, we account for real-world uncertainties, including odometry errors, actuator
imperfections, and orientation sensor inaccuracies. In practice, the direction $\mathbf{d}_k$ in the navigation frame
$(\mathbf{e}_x,\mathbf{e}_y)$ is subject to orientation errors, and the distance step
$v\,\Delta t$ may not be executed exactly as commanded due to external or internal perturbations. The dynamic equation for the position is then given by

\begin{equation}
\label{eq:dynamic_pos}
 \mathbf{p}_{k+1} 
= \mathbf{p}_{k} 
+ f\!\left(
    v\, \Delta t\,\mathbf{d}_k, \mathbf{n}_{k}
\right)
\end{equation}

with 
\[
\mathbf{d}_k = \frac{ \tilde{\mathbf{G}}_{k}^{\mathsf T} \mathbf{N}^{-1} \bigl( \mathbf{B}_g - \mathbf{B}_{k} \bigr) }{ \left\| \tilde{\mathbf{G}}_{k}^{\mathsf T} \mathbf{N}^{-1}\bigl( \mathbf{B}_g - \mathbf{B}_{k} \bigr) \right\| },
\]         
where $\mathbf{n}_{k}$ is a random variable drawn from some distribution (not necessarily normal) that models  the  imperfections of the control command, so that
\[
f\!\left(
    v\, \Delta t\,\mathbf{d}_k, \mathbf{n}_{k}
\right)
\]
represents the noisy version of the desired control command \(v\, \Delta t\,\mathbf{d}_k\). In general, the form of the noisy control depends on the kinematics of the system and is nonlinear, denoted by $f(\cdot)$.
In this paper, this is modeled by perturbing the heading angle, denoted by $\theta_k$, and the traveled distance.
The heading angle $\theta_k$ is affected by additive white noise $\eta_k$ with standard deviation $\delta_\theta$,
while the traveled distance $v\,\Delta t$ is affected by additive white noise $\xi_k$ with standard deviation $\delta_d$. Consequently, if the desired control is
\[
v\,\Delta t\,\mathbf{d}_k
= v\,\Delta t \left( \cos(\theta_k)\mathbf{e}_x + \sin(\theta_k)\mathbf{e}_y \right),
\]
the noisy control $f\!\left(
    v\, \Delta t\,\mathbf{d}_k, \mathbf{n}_{k}
\right)$ becomes
\begin{equation}
    \label{eq:noisy_control}
    (v\,\Delta t + \xi_k)
\left( \cos(\theta_k + \eta_k)\mathbf{e}_x + \sin(\theta_k + \eta_k)\mathbf{e}_y \right),
\end{equation}

where $\xi_k \sim \mathcal{N}(0,\delta_d^2)$ and $\eta_k \sim \mathcal{N}(0,\delta_\theta^2)$.


\subsection{Rao--Blackwellized Particle Filter}
\label{subsec:rb_filter}
We denote by $\big( \mathbf{p}_{k}, \mathbf{g}_{k} \big)$ the state of the system~(\ref{eq:system}), defined by the two dynamic equations~(\ref{eq:dynamic_pos}), (\ref{eq:dyn_gradient}), and the observation equation~(\ref{eq:observation_v1}):

\begin{equation}
\label{eq:system}
\left\{
\begin{array}{l}
\mathbf{p}_{k+1} 
= \mathbf{p}_{k} 
+ f\!\left(
    v\, \Delta t\,
    \frac{ \tilde{\mathbf{G}}_{k}^{\mathsf T} \mathbf{N}^{-1}\bigl( \mathbf{B}_g - \mathbf{B}_{k} \bigr) }
         { \left\| \tilde{\mathbf{G}}_{k}^{\mathsf T} \mathbf{N}^{-1}\bigl( \mathbf{B}_g - \mathbf{B}_{k} \bigr) \right\| },
    \mathbf{n}_{k}
\right), \\[1.2em]

\mathbf{g}_{k+1} 
= \mathbf{g}_{k} + \mathbf{q}_{k}, \\[1.0em]

\mathbf{B}_{k+1} 
= \mathbf{B}_{k} + \mathbf{H}_{k}\, \mathbf{g}_{k} + \mathbf{a}_k + \mathbf{r}_{k+1}.
\end{array}
\right.
\end{equation}

At the initial time $t_0$ (starting point), we assume that $\tilde{\mathbf{G}}_{0}$ is approximately known, with a certain uncertainty, from a prior distribution. This initial estimate is used to perform the first gradient descent step in~\eqref{eq:dynamic_pos}. At the new position, the gradient is re-estimated by considering its expected variation in~\eqref{eq:dyn_gradient}, together with the new measurements and the corresponding predicted measurements in ~\eqref{eq:observation_v1}. A new gradient descent step is then computed, and the process is repeated iteratively.

The remaining question is: how can we obtain a reliable estimate of the state of system~(\ref{eq:system}), which we denote by $\big(\tilde{\mathbf{p}}_{k}, \tilde{\mathbf{g}}_{k}\big)$? In estimation theory, when the system is linear with respect to the state and subject to additive, independent, zero-mean noise, the well-known Kalman filter provides the online minimum mean-squared error (MMSE) estimate~\cite{humpherys_fresh_2012, uhlmann_gaussianity_2022}.

The system~(\ref{eq:system}) contains two nonlinear equations~(\ref{eq:dynamic_pos}) and (\ref{eq:observation_v1}). The nonlinearity arises from two sources. First, the term $\mathbf{H}_{k}\mathbf{g}_{k}$ represents a multiplication between components of the state vector. 
Second, the term
\[
\frac{ \tilde{\mathbf{G}}_{k}^{\mathsf T} \mathbf{N}^{-1}\bigl( \mathbf{B}_g - \mathbf{B}_{k} \bigr) }
{ \left\| \tilde{\mathbf{G}}_{k}^{\mathsf T} \mathbf{N}^{-1}\bigl( \mathbf{B}_g - \mathbf{B}_{k} \bigr) \right\| }
\]
is nonlinear due to the normalization operation. This makes the direct use of a Kalman filter (KF) suboptimal. However, the joint probability density function of the state 
$\big(\mathbf{p}_{k}, \mathbf{g}_{k}\big)$ 
can be decomposed using Bayes' rule as

\[
p(\mathbf{p}_{k}, \mathbf{g}_{k} \mid \mathbf{B}_k) 
= p(\mathbf{g}_{k} \mid \mathbf{p}_{k}, \mathbf{B}_k)\,
  p(\mathbf{p}_{k} \mid \mathbf{B}_k).
\]

The term $p(\mathbf{g}_{k} \mid \mathbf{p}_{k}, \mathbf{B}_k)$ is conditionally linear; 
that is, in equation~(\ref{eq:observation_v1}), $\mathbf{B}_{k+1}$ becomes linear with respect to $\mathbf{g}_{k}$, 
given the probability mass function of $\mathbf{p}_{k}$ (knowing the matrix $\mathbf{H}_{k}$, which depends on $\mathbf{p}_{k}$). 
This implies that the gradient $\mathbf{g}_{k}$ can be estimated using a KF. The second term, $p(\mathbf{p}_{k} \mid \mathbf{B}_k)$, can be estimated using nonlinear filtering techniques such as Sequential Monte Carlo (SMC) particle filtering, where the probability density function is approximated by a weighted and normalized sum of $N_{p}$ particles: 
\[
p(\mathbf{p}_{k} \mid \mathbf{B}_k) \approx \sum_{i=1}^{N_{p}} w^{i}_k \, \delta_{\mathbf{p}^{i}_{k}}(\mathbf{p}_{k}),
\]

where $\delta_{\mathbf{p}^{i}_{k}}(\mathbf{p}_{k})$ is the Dirac measure centered at particle $\mathbf{p}^{i}_{k}$, defined as
\[
\delta_{\mathbf{p}^{i}_{k}}(\mathbf{p}_{k}) =
\left\{
\begin{array}{ll}
1, & \mathrm{if}\ \mathbf{p}_{k}=\mathbf{p}^{i}_{k}, \\[6pt]
0, & \mathrm{otherwise}
\end{array}
\right.
\] and the particle weights $w^{i}_k \geq 0$ are normalized such that
\[
\sum_{i=1}^{N_{p}} w^{i}_k = 1.
\]

The integration of a particle filter to estimate the nonlinear part of the system state (the position in our case), together with a Kalman filter to estimate the conditionally linear substate (the gradient in our case), is known as the RBPF, or equivalently, a mixture Kalman filter \cite{hendeby_rao-blackwellized_2010}.

The RBPF is a filtering method that lies between two extremes. On one side, a full particle filter estimates the entire state $\big(\mathbf{p}_{k}, \mathbf{g}_{k}\big)$, which has dimension \((2n+2)\). Although this approach is very general and can theoretically handle all types of nonlinearities, it requires a very large number of particles to achieve sufficient density in such a high-dimensional space. In practice, the number of particles is constrained by memory, computational resources, and real-time requirements. On the other side, linearizing the system with an Extended Kalman Filter (EKF) to completely avoid particle filtering may introduce additional approximation errors into the already approximated system~(\ref{eq:system}).

The RBPF offers a compromise between these two methods. In our case, the particle filter is applied only to the low-dimensional component (position, \(\mathbf{p}_{k}\), which is a 2-dimensional vector), thereby avoiding the curse of dimensionality, while the Kalman filter is used for the higher-dimensional component (gradient, \(\mathbf{g}_{k}\), which is of dimension $2n$).

To apply RBPF, we first initialize, at time step $t_0$, the particle positions $(\mathbf{p}^{i}_0)_{1 \leq i \leq N_p}$ around the starting location with a certain resolution in the navigation frame, along with their corresponding weights $(w^{i}_0)_{1 \leq i \leq N_p}$. The gradient $\mathbf{g}_0$ at the starting location is supposed to be approximately known, 
$\mathbf{g}_0 \sim \mathcal{N}(\tilde{\mathbf{g}}_0, \mathbf{C}_{0})$.
The update steps are summarized in the pseudo code (Algorithm~\ref{algo:rbpf_gd}).

\begin{algorithm}[t]
\caption{RBPF Gradient Descent for BIMN}
\label{algo:rbpf_gd}
\begin{algorithmic}[1]
\ForAll{particles $i$}
  \State $\mathbf{p}^{i}_{k+1}
  = \mathbf{p}^{i}_{k}
  + f\!\left(
      v\, \Delta t\,
      \frac{ \tilde{\mathbf{G}}_{k}^{\mathsf T}\mathbf{N}^{-1}\bigl( \mathbf{B}_g - \mathbf{B}_{k} \bigr) }
           { \left\| \tilde{\mathbf{G}}_{k}^{\mathsf T}\mathbf{N}^{-1}\bigl( \mathbf{B}_g - \mathbf{B}_{k} \bigr) \right\| },
      \mathbf{n}^i_{k}
    \right)$ \Comment{Gradient descent step}

  \State $\mathbf{C}^{i}_{k+1} = \mathbf{C}^{i}_{k} + \mathbf{Q}_{k}$ \Comment{Gradient covariance update}

  \State $\boldsymbol{\Sigma}^i_{k+1}
  = \mathbf{R}_k + \mathbf{H}_k^i \mathbf{C}^{i}_{k+1} (\mathbf{H}_k^i)^{\mathsf T} + \mathbf{A}_k + \mathbf{R}_{k+1}$

  \State $\mathbf{B}^i_{k+1}
  = \mathbf{B}_{k} + \mathbf{H}^{i}_{k}\, \mathbf{g}^{i}_{k}$ \Comment{Measurement prediction}

  \State $\tilde w^i_{k+1}
  = w^i_{k}\cdot \mathcal{N}\!\bigl(\mathbf{B}_{k+1}-\mathbf{B}^{i}_{k+1},\, \boldsymbol{\Sigma}^i_{k+1}\bigr)$ \Comment{Weight correction using the measurement}

  \State $\mathbf{K}^i_{k+1}
  = \mathbf{C}^i_{k+1} (\mathbf{H}_k^i)^{\mathsf T}\, (\boldsymbol{\Sigma}^i_{k+1})^{-1}$

  \State $\mathbf{g}_{k+1}^i
  = \mathbf{g}_{k}^i + \mathbf{K}^i_{k+1} \left(\mathbf{B}_{k+1}-\mathbf{B}^{i}_{k+1}\right)$ \Comment{Kalman gradient correction}

  \State $\mathbf{C}^i_{k+1}
  = \mathbf{C}^i_{k+1} - \mathbf{K}^i_{k+1} \boldsymbol{\Sigma}^i_{k+1} (\mathbf{K}^i_{k+1})^{\mathsf T}$ \Comment{Kalman covariance correction}
\EndFor

\ForAll{particles $i$}
  \State $w^i_{k+1} = \tilde w^i_{k+1} \Big/ \sum_{j=1}^{N_p} \tilde w^j_{k+1}$ \Comment{Weights normalization}
\EndFor

\State $\tilde{\mathbf{g}}_{k+1} = \sum_{i=1}^{N_p} w^i_{k+1}\,\mathbf{g}_{k+1}^{i}$ \Comment{Estimated gradient for the descent step}
\end{algorithmic}
\end{algorithm}

In addition, the position of the agent in the navigation frame can be estimated as
\begin{equation}
\label{eq:estimated_postion_by_mean}
    \tilde{\mathbf{p}}_{k+1} = \sum_{i=1}^{N_p} w_{k+1}^i \, \mathbf{p}_{k+1}^i .
\end{equation}

The uncertainty covariance matrices of the estimated position (denoted $\tilde{\mathbf{P}}_{k+1}$) 
and gradient (denoted $\tilde{\mathbf{C}}_{k+1}$) are given by
\[
\tilde{\mathbf{P}}_{k+1} 
= \sum_{i=1}^{N_p} w_{k+1}^i 
\, (\mathbf{p}^i_{k+1}-\tilde{\mathbf{p}}_{k+1})^{\mathsf T}
(\mathbf{p}^i_{k+1}-\tilde{\mathbf{p}}_{k+1}),
\]
\[
\tilde{\mathbf{C}}_{k+1} 
= \sum_{i=1}^{N_p} w_{k+1}^i 
\left[ \mathbf{C}^{i}_{k+1} 
+ (\mathbf{g}^i_{k+1}-\tilde{\mathbf{g}}_{k+1})^{\mathsf T}
(\mathbf{g}^i_{k+1}-\tilde{\mathbf{g}}_{k+1}) \right].
\]

The process of weight correction (Algorithm~\ref{algo:rbpf_gd}, line 6) can lead to degeneracy; some of the major particles may end up with very small weights. These particles do not contribute to decision making and only slow down the computation. This is why re-sampling is required from time to time to maintain an effective particle distribution \cite{li_resampling_2015}. The most common criterion for re-sampling is the \emph{effective sample size (ESS)} \cite{martino_effective_2017}, where resampling is performed whenever
\[
N_{\mathrm{eff}} = \frac{1}{\sum_{i=1}^{N_p} w_i^2} \leq \theta_{\mathrm{eff}} N_p,
\]
after which all particles are assigned equal weights. Here, $\theta_{\mathrm{eff}} \in [0,1]$ is a threshold parameter, typically set to $0.5$, meaning that re-sampling is triggered when the effective sample size $N_{\mathrm{eff}}$ falls below half of the total number of particles $N_p$. Weight correction can cause weights to become extremely small and fall below machine precision, due to the likelihood term \(\mathcal{N}\!\bigl(\mathbf{B}_{k+1}-\mathbf{B}^{i}_{k+1},\, \boldsymbol{\Sigma}^i_{k+1}\bigr)\), which involves exponentials of negative values. To address this issue, particle weights are computed in the logarithmic domain, which improves numerical stability~\cite{gentner_log-pf_2018}. This log-domain computation is implemented but not discussed here.

The proposed method is capable of estimating the gradient, which is then used to guide the agent toward the goal location by minimizing the objective function. However, it inherits the main weakness of gradient descent: it can become trapped in a local extremum where the gradient is zero. Nevertheless, the agent includes a noise model that may help it overcome this difficulty.

\section{RESULTS}
\label{sec:results}

The proposed method is tested in simulation using the fourteenth generation of the International Geomagnetic Reference Field (IGRF-14) \cite{IAGA_IGRF14_2024_v2}. The IGRF is a standard mathematical model that describes the Earth’s main magnetic field. It is derived from real data collected from satellites, observatories, and surveys around the globe and is maintained and updated every five years.

\begin{figure}[!t]
    \centering
    \includegraphics[width=0.8\columnwidth]{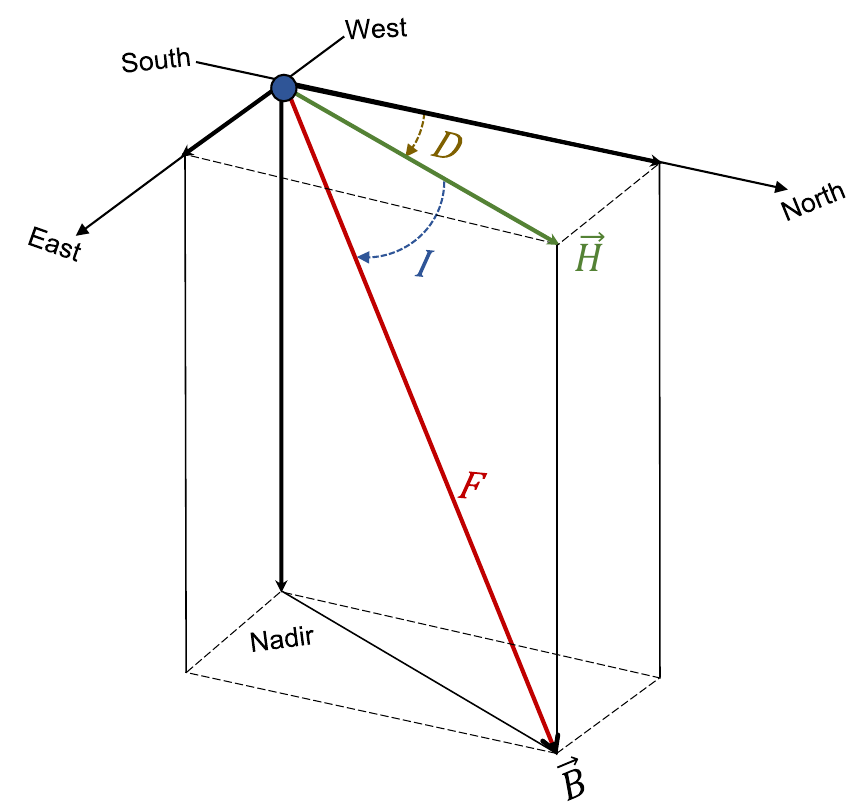}
    \caption{In the (north, east, nadir) reference frame, the inclination ($I$) is the angle between the magnetic field vector $\vec{B}$ and its horizontal projection $\vec{H}$. The declination ($D$) is the angle between $\vec{H}$ and north.}
    \label{fig:inclination_declination_intensity}
\end{figure}

The Earth magnetic field can be characterized by three magnetic parameters: the inclination ($I$), declination ($D$), and intensity ($F$) (see Fig.~\ref{fig:inclination_declination_intensity}). These parameters provide useful spatial variations that can be used in navigation. In practice, these parameters can be obtained using a magnetometer that measures the magnetic field vector. The horizontal plane required for inclination estimation is determined from accelerometer measurements that provide the local vertical, while geographic north for declination estimation is obtained using a gyroscope, a sun sensor, or an external reference such as a star-tracking camera.

\begin{figure*}[t]
    \centering
    \includegraphics[width=1.0\textwidth]{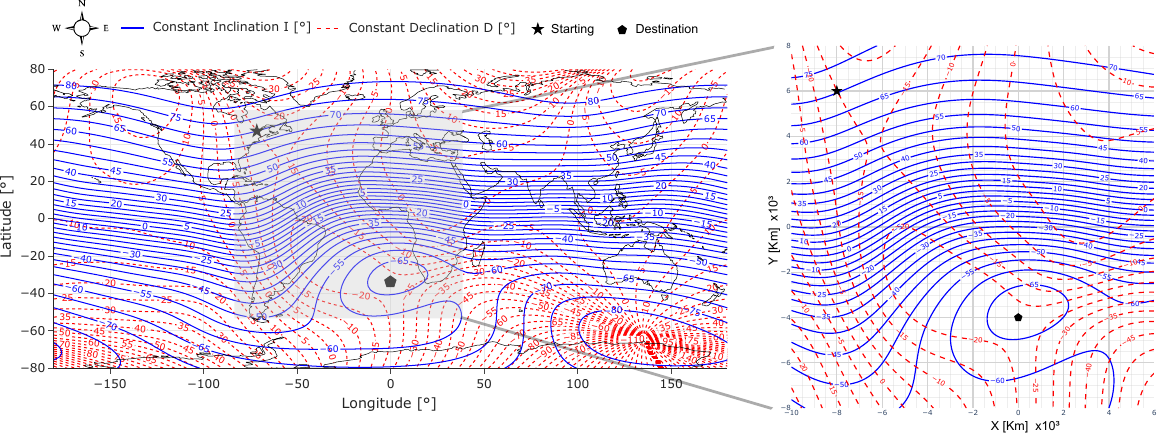}
    \caption{Global Mercator projection of geomagnetic inclination and declination derived from the IGRF-14 model at sea level on 1 January 2025. The experimental area is shown in gray and represented in a two-dimensional Cartesian coordinate system. The goal is to navigate from the start point ($-71.9^\circ$, $47.3^\circ$), corresponding to $(-8,\,6)\times10^{3}$~km, to the destination ($0.0^\circ$, $-34.0^\circ$), corresponding to $(0,\,-4)\times10^{3}$~km. The destination is chosen because it has unique magnetic parameters $(I, D)=(-67.23^\circ, -21.79^\circ)$
    within the experimental area, and no map information is available to the agent; only the magnetic parameters of the destination and the currently measured magnetic parameters are known. The navigation frame is centered at the starting position, with the $x$-axis pointing east and the $y$-axis pointing north.}
    \label{fig:zone_of_experiment}
\end{figure*}

We perform two experiments. The first experiment (see \ref{subsec:ideal_conditions}) considers an ideal case in which measurement noise and real-world platform drift are not included. The goal of this experiment is to demonstrate that the proposed method continuously estimates the gradient in real time and uses it to guide the system toward the destination. We also compare our approach with the reference method \cite{gill_navigation_2024} that assumes a constant gradient, in order to assess whether estimating the gradient is beneficial. The second experiment (see \ref{subsec:noise_drift}) is more realistic and reflects real-world conditions, where measurement noise and drift are simulated. Measurement noise arises from the fact that the sensor and its calibration are never perfect, while drift results from the system not executing the commands perfectly in both the heading and the distance step.

In the following two experiments, only the magnetic inclination $I$ and declination $D$ are used. 
These parameters are chosen for two main reasons. 
First, they are the most commonly used quantities in early and classical compass-based marine navigation. 
Second, inclination and declination vary smoothly and predictably over the Earth’s surface 
and are homogeneous quantities expressed in degrees. However, in practice, the magnetometer measures in its body-fixed frame rather than the (north, east, nadir) reference frame. The inclination $I$ and the intensity $F$ are therefore more suitable for real-world applications, as they are easier to extract than the declination.

For both experiments, we use the Mercator projection, which is the standard map projection 
used in classical compass-based marine navigation. In this representation, the Earth’s surface is projected onto a two-dimensional plane. The Mercator projection is conformal, so directions are preserved, making the projection well suited for navigation: 
a local direction on the map corresponds to the same direction on the Earth’s surface.

The map frame is centered at latitude $0^\circ$ and longitude $0^\circ$. 
The horizontal axis corresponds to the west--east direction, and the vertical axis 
corresponds to the south--north direction.

The conversion from geographic coordinates (latitude $\phi$, longitude $\lambda$) 
to Mercator coordinates $(x,y)$ is given by
\[
\left\{
\begin{aligned}
x &= R \lambda, \\
y &= R \ln \left( \tan\left( \frac{\pi}{4} + \frac{\phi}{2} \right) \right),
\end{aligned}
\right.
\]
where $R$ is the Earth’s radius, $\lambda$ is expressed in radians, and $\phi$ denotes 
the latitude. For the map and the area of the experiment, see Fig.~\ref{fig:zone_of_experiment}. The RBPF method is initialized with  velocity $v = 100~\mathrm{km/h}$ and sampling period $\Delta t = 0.05~\mathrm{h}$, which corresponds to a distance step of $5~\mathrm{km}$. At the initial position $\mathbf{p}_0$, it is also initialized with the true local gradient, 
\[
\mathbf{G}_0 =
\begin{bmatrix}
-0.002 & \;\;0.004 \\
-0.003 & -0.002
\end{bmatrix}
\ \mathrm{[^{\circ}/km]}.
\]

The initial uncertainty of the given gradient $\mathbf{G}_0$ is set to
\[
\mathbf{C}_0 = (0.005)^2 \mathbf{I}_4,
\]
where $\mathbf{I}_4$ is the $4 \times 4$ identity matrix. This corresponds to a standard deviation of $0.005$~°/km, indicating that the true gradient is expected to lie within this uncertainty range. The uncertainty of the gradient variation in~\eqref{eq:dyn_gradient} is set to a constant value at each step:
\[
\mathbf{Q}_k = (0.001)^2 \mathbf{I}_4 .
\]

This means that the gradient after a step distance of $5~\mathrm{km}$ is assumed to be equal to the previous one (slowly varying gradient), with an increasing uncertainty corresponding to a standard deviation of $0.001~^\circ/\mathrm{km}$, which is of the same order as the global gradient variation. To account for approximation errors in the observation equation~\eqref{eq:observation_v1}, the covariance matrix $\mathbf{A}_k$ is set to a constant value:
\[
\mathbf{A}_k = (0.005)^2 \mathbf{I}_2 .
\]
This corresponds to a standard deviation of $0.005^\circ$.  
It comes from the assumed gradient uncertainty of $0.001^\circ/\mathrm{km}$.  
Since the observation model contains the term $\mathbf{H}_{k}\mathbf{g}_{k}$, over a distance step of $5~\mathrm{km}$ this leads to approximately a standard deviation of
$5 \times 0.001 = 0.005^\circ$ in the observations.
The number of particles is $2{,}500$ ($50 \times 50$), initialized with equal weights within a $10 \times 10~\mathrm{km}$ square centered at the starting position.

The stopping threshold is set to $\epsilon = 0.1^\circ$ for both experiments. The destination zone, defined by $\|\mathbf{B}_g - \mathbf{B}_{k_0}\| \leq \epsilon$, is shown in light gray in Fig.~\ref{fig:compared_two_trajectories}. This region may limit the accuracy of reaching the destination. Its size can be reduced by decreasing $\epsilon$, adjusting the normalization parameters $\sigma_1$ and $\sigma_2$ (set to $1.0$ in both simulations), or by incorporating additional magnetic parameters.

The destination is located in a region where simple strategies such as following an isoline of constant value or moving in the globally decreasing direction of inclination or declination are ineffective. In contrast, the following simulations show that the gradient descent method naturally guides the trajectory toward the destination and remains robust to heading and orientation errors, since the direction decision depends only on the current gradient estimate and not on past decisions. In other words, the destination acts as an attraction that pulls the gradient descent method toward it.

\subsection{Experiment under ideal conditions}
\label{subsec:ideal_conditions}

 In this experiment, no measurement noise is simulated, so $\mathbf{R}_k$ is set to zero. At each step, the particles are perturbed by $\delta_d = 10\,\mathrm{m}$ in distance and
$\delta_\theta = 0.5^\circ$ in heading, using \eqref{eq:noisy_control}, to maintain state-space exploration and to explore the nonlinearity in \eqref{eq:dynamic_pos}, even though command imperfections are not simulated in this experiment. The reference method we compare against, which assumes a constant gradient, is initialized using the gradient value at the initial position $\mathbf{G}_0$, with the same velocity and sampling period as the RBPF method.

\begin{figure}[!t]
    \centering
    \includegraphics[width=0.9\columnwidth]{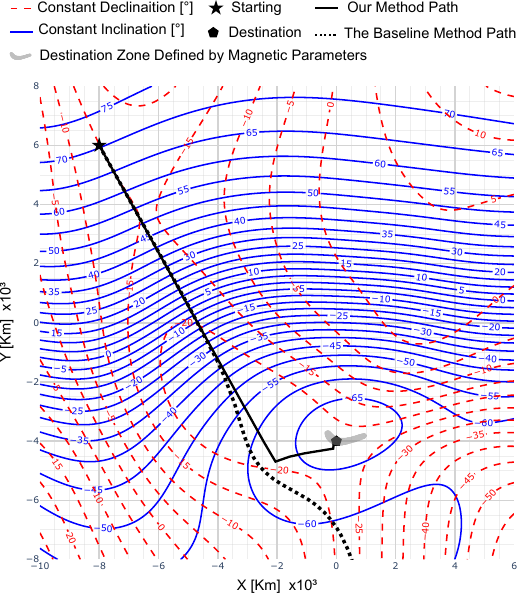}
    \caption{The navigation paths of both methods show that the proposed method is attracted to the destination, whereas the baseline method diverges.
}
    \label{fig:compared_two_trajectories}
\end{figure}

\begin{figure}[!htbp]
    \centering
    \includegraphics[width=0.86\columnwidth]{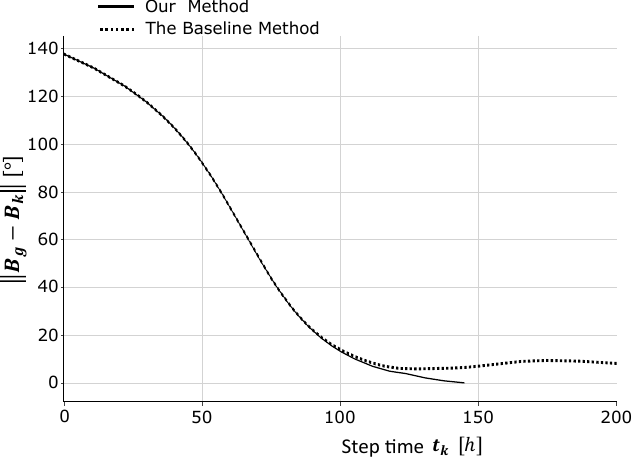}
    \caption{Variation of the objective function for the two methods. The proposed method exhibits a monotonic decrease due to gradient descent, whereas the baseline method does not, particularly near the end of the trajectory where the initial gradient no longer matches the local gradient.}
    \label{fig:Losses}
\end{figure}

The comparison between the two methods shows that, since both are initialized at the same starting position with the same initial gradient, they initially follow the same trajectory. However, the baseline method eventually diverges because it does not incorporate an estimation of the local gradient, unlike the proposed method (see Figs.~\ref{fig:compared_two_trajectories} and~\ref{fig:Losses} for the navigation paths of the two methods and the corresponding variations of the objective function). The baseline method is suitable for local navigation in regions where the gradient can be assumed constant, so there is no need for additional estimation complexity. For large-scale navigation, this assumption no longer holds, and gradient estimation becomes necessary.

\begin{figure*}[t]
    \centering
    \includegraphics[width=1.7\columnwidth]{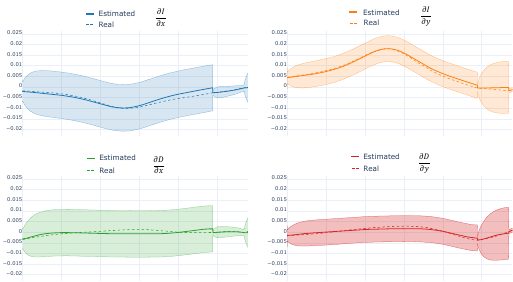}
    \caption{ This figure shows the estimated gradient with respect to the true gradient, along with the uncertainty region corresponding to one standard deviation. Uncertainty decreases approximately along the direction of motion. This can be seen clearly in the two abrupt changes in uncertainty,
which correspond to changes in direction approximately along
$\mathbf{e}_x$ and then $\mathbf{e}_y$. The $y$-axis is expressed in $^\circ/\mathrm{km}$, and the $x$-axis (not shown) represents time.}
    \label{fig:gradient_estimation}
\end{figure*}

\begin{figure}[!htbp]
    \centering
    \includegraphics[width=0.9\columnwidth]{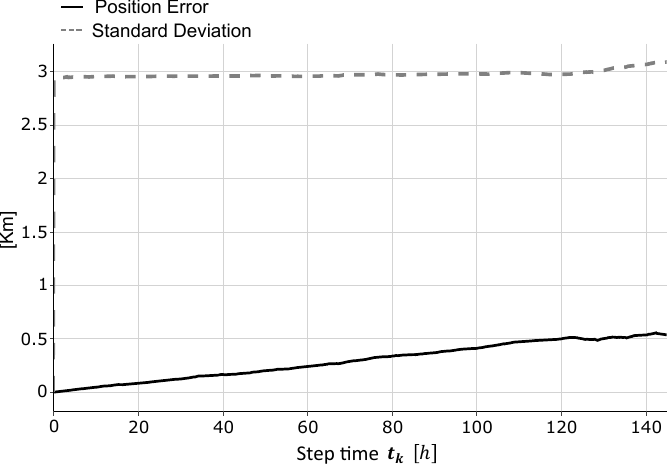}
    \caption{For the RBPF method, the error between the estimated and true positions reaches a maximum of $500\,\mathrm{m}$ at the end of the trajectory, for a traveled distance of more than $12\,000\,\mathrm{km}$.}
    \label{fig:position_error_estimated}
\end{figure}

For the proposed method, the estimated local gradient is shown in Fig.~\ref{fig:gradient_estimation}. The figure shows that the gradient is estimated along the path with good accuracy, up to an uncertainty standard deviation corresponding to the order of the gradient variation. The position estimation error, which corresponds to the norm of the difference between
the estimated position given by \eqref{eq:estimated_postion_by_mean} and the ground-truth
position, is shown in Fig.~\ref{fig:position_error_estimated} and indicates a good position estimate with an uncertainty of about $3\,\mathrm{km}$. This position error standard deviation is due to the fact that the magnetic parameters are almost constant at the $1\,\mathrm{km}$ scale. In other words, the slow and smooth variation of the magnetic parameters makes it difficult to accurately constrain the particle weights in the exact position.

\subsection{Experiment under noise and drift}
\label{subsec:noise_drift}

For this experiment, measurement noise on both magnetic parameters, $I$ and $D$, is modeled as white Gaussian noise with a standard deviation of $0.05^\circ$. To provide an order-of-magnitude interpretation, consider a mean Earth's magnetic field strength of $45\,\mu\text{T}$. The corresponding error magnitude $\xi$ can be approximated using
\[
\tan(0.05^\circ) \approx \frac{\xi}{45\,\mu\text{T}},
\]
which yields $\xi \approx 40\,\text{nT}$. Thus, an angular noise level of $0.05^\circ$ corresponds to a transverse magnetic field error of approximately $40\,\text{nT}$ for a nominal field strength of $45\,\mu\text{T}$, which is comparable to typical low-cost magnetometer noise. The system command imperfections, which induce drift, are modeled in both the distance and heading commands as white Gaussian noise, with standard deviations $\delta_d = 1\,\mathrm{km}$ and $\delta_\theta = 15^\circ$, as described in \eqref{eq:noisy_control}.

In the RBPF method, the same noise levels are used to perturb the particles at each step. Each particle is therefore perturbed by sampled from $\mathcal{N}(0,\delta_d^2)$ for distance and $\mathcal{N}(0,\delta_\theta^2)$ for heading. This allows the particles to explore the same drift uncertainty as the simulated system. To avoid particle degeneracy and overconcentration around the true position, the measurement noise matrix is set to $\mathbf{R}_k = (0.2)^2 \mathbf{I}_2$. Although the simulated measurement noise is $0.05^\circ$, a larger value of $0.2^\circ$ is used to prevent overconfidence and to account for model uncertainties, thereby improving particle diversity and stability.

\begin{figure*}[t]
    \centering
    \includegraphics[width=1.7\columnwidth]{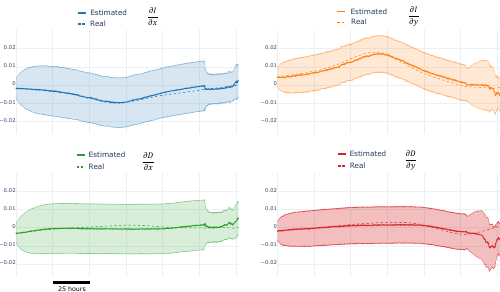}
    \caption{The figure shows the evolution of the estimated gradient and the true gradient in the presence of noise and drift. The uncertainty matches the observation noise level $\mathbf{R}_k = (0.2)^2 \mathbf{I}_2$. Toward the end, the true gradient approaches zero, making it difficult to observe given this level of noise.}
    \label{fig:gradient_estimation_with_noise}
\end{figure*}

\begin{figure}[!htbp]
    \centering
    \includegraphics[width=\columnwidth]{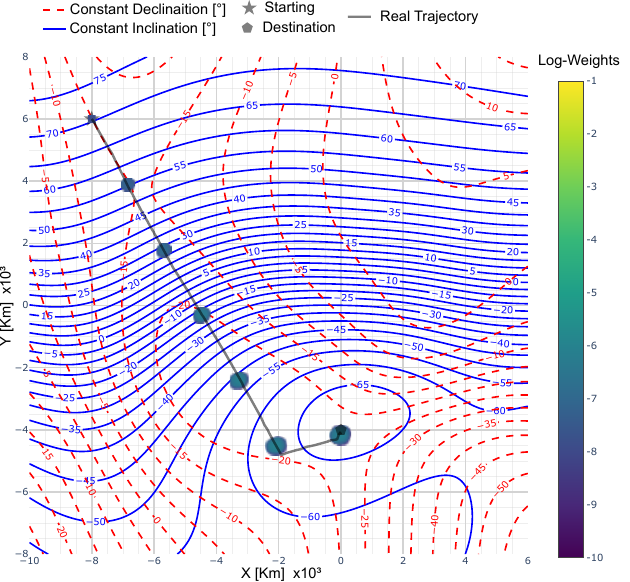}
    \caption{The figure shows the distribution of particles at different time steps. For clarity, the distribution is zoomed in by a factor of 5/3. The particles remain around the true position and form a local, weighted map of possible positions. The evolution of the distribution suggests a unimodal behavior, in which centrally located particles carry higher weights than those at the periphery, indicating that an EKF may adequately capture this evolution. We also note that no particle resampling occurred during this experiment with $\theta_{\mathrm{eff}} = 0.5$, which means that more than half of the particles remain effective.}
    \label{fig:Particles_evolution}
\end{figure}

\begin{figure}[!htbp]
    \centering
    \includegraphics[width=\columnwidth]{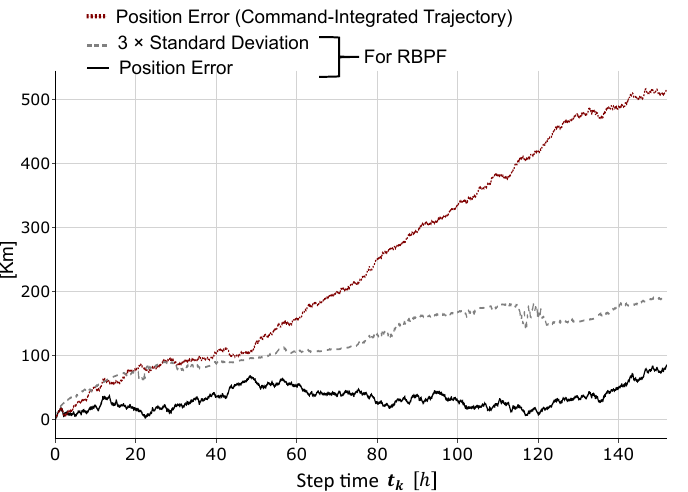}
    \caption{The figure clearly shows the attenuation of drift when using the estimated gradient for localization in the proposed method. We can see that the error sometimes increases and then decreases; for example, around 120 h, accurate relocalization is achieved (approximately 10 km). For the RBPF, drift is observed toward the end, likely due to poor gradient estimation as the system approaches the destination (see Fig.~\ref{fig:gradient_estimation_with_noise}). This indicates that the proposed method provides relocalization in addition to guiding the system toward the destination. In contrast, when the commands are blindly integrated, the drift increases linearly and exceeds 500 km.}
    \label{fig:position_error_estimated_vs_integrated}
\end{figure}

The experiment provides insight into the robustness of the proposed method against potential real-world perturbations. Despite the presence of noise, the gradient is well estimated (see Fig.~\ref{fig:gradient_estimation_with_noise}) and effectively guides the system toward the destination (see Fig.\ref{fig:Particles_evolution}). Near the end of the trajectory, as the system approaches the destination, the gradient becomes poorly estimated. This is because it nearly vanishes in the destination region, and the presence of noise makes it difficult to observe. In addition to estimating the gradient, the method also estimates the current position of the system in the navigation frame, providing a form of localization alongside the guidance component. To assess whether the proposed method improves position estimation, we compare the RBPF estimate with the naive trajectory obtained by purely integrating the gradient descent commands, assuming perfect execution of the commands. The purely integrated command trajectory diverges from the real position due to the simulated drift, which prevents the system from executing the exact command. On the other hand, the proposed method makes use of the locally estimated gradient to attenuate the drift of the estimated trajectory (see Fig.~\ref{fig:position_error_estimated_vs_integrated}). Through gradient estimation, the RBPF method implicitly constructs a local map around the true position, which is then used for system relocalization.



The proposed method guides the system through gradient estimation and does not depend on the previously integrated position (see Eq.~\ref{eq:observation_v0}), as the observation equation depends only on the current position change ($\mathbf{p}_{k+1} - \mathbf{p}_{k}$). In contrast, drift in the purely integrated command trajectory persists and influences future decisions, since the current position contains all accumulated drift from the past. The RBPF method is well suited for guidance; however, its localization accuracy may be affected by weak gradient variations. On the other hand, if the gradient is locally strong and accurately estimated, the weighted distribution of possible positions can concentrate around the true position. That means traversing high-information regions (i.e., regions with strong gradients) can improve localization performance. This is why planning trajectories that pass through such regions is beneficial for relocalization \cite{penumarti_real-time_2026}. A trade-off remains between the ability to estimate the gradient, which requires it to vary slowly, and the presence of strongly varying gradients that are more difficult to track.

\section{CONCLUSIONS AND FUTURE WORK}
This paper presents a gradient-descent–based method for BIMN using RBPF to locally estimate the gradient and guide the system toward the destination. The simulation results indicate that the proposed approach remains effective for navigation in magnetic environments, even when the gradient varies. This type of method has the advantage of not relying on any pre-constructed magnetic map, making it potentially lightweight, although its computational cost depends on the particle filter resampling rate. The method relies on several parameters that must be carefully tuned before the navigation mission. These parameters should take into account the command imperfections, measurement noise, and the complexity of the magnetic environment. In addition, the distance step should be optimally chosen to ensure convergence.

A key limitation of BIMN, not addressed in this paper, is that multiple spatial locations may share identical magnetic parameters. This can cause BIMN to converge to an incorrect location. One way to mitigate this issue is to incorporate additional parameters. However, a key challenge is that the magnetic field provides at most three parameters (intensity, inclination, and declination), potentially limiting the uniqueness of these parameters. Therefore, the approach can be extended by including other quantities, such as the gravity field or pressure, to improve the uniqueness of the signature associated with a given destination. Alternatively, without increasing the number of parameters, intermediate waypoints can be introduced. The sequence of waypoints can be designed to guide BIMN and enforce convergence to the true destination, while forcing the system to pass through high-information areas to improve localization. This waypoint-based strategy will be investigated in future work.
Future work will also focus on studying a linearized version of the proposed method using an EKF and evaluating whether its performance is acceptable compared to the full nonlinear formulation. In addition, this study assumed time-invariant magnetic parameters. In practice, external disturbances such as solar activity can perturb the Earth’s magnetic field. Therefore, further investigation is required to see if the method is robust to time-varying magnetic conditions under such perturbations. Finally, the method will be evaluated in more complex magnetic environments, where variations of magnetic parameters are not smooth, such as indoor environments.
\section{ACKNOWLEDGMENTS}
We thank the Direction Générale de l’Armement (DGA) and the Agence pour l’Innovation de Défense (AID) for technical and scientific supporting and partial funding of this work. We also thank Nathanaël Schaeffer for his helpful feedback in improving the quality of this paper.

\bibliographystyle{IEEEtran}
\bibliography{references}

\end{document}